\documentclass[10pt,conference]{IEEEtran}\IEEEoverridecommandlockouts

\usepackage{amsmath,amssymb,amsfonts}
\usepackage{graphicx}
\usepackage{textcomp}
\usepackage{xcolor}
\usepackage{multirow}
\usepackage{booktabs}
\usepackage{balance}
\usepackage[ruled, linesnumbered, noline]{algorithm2e}
\usepackage{mathtools}

\DeclarePairedDelimiter\ceil{\lceil}{\rceil}
\DeclarePairedDelimiter\floor{\lfloor}{\rfloor}

 \usepackage[
    backend=biber,
    style=ieee,
  ]{biblatex}
\begin{document}

\title{SCAT: Second Chance Autoencoder for Textual Data}

\title{\huge{\textbf{SCAT: Second Chance Autoencoder for Textual Data}}}

\author{\IEEEauthorblockN{Somaieh Goudarzvand, Gharib Gharibi, Yugyung Lee} \vspace{1mm}
\IEEEauthorblockA{\textit{School of Computing and Engineering, University of Missouri-Kansas City} 
Kansas City, USA\\ 
\{sgnbx, ggk89\}@mail.umkc.edu, LeeYu@umkc.edu}}

\maketitle

\begin{abstract}
We present a k-competitive learning approach for textual autoencoders named Second Chance Autoencoder (SCAT). SCAT selects the $k$ largest and smallest positive activations as the winner neurons, which gain the activation values of the loser neurons during the learning process, and thus focus on retrieving well-representative features for topics. Our experiments show that SCAT achieves outstanding performance in classification, topic modeling, and document visualization compared to LDA, K-Sparse, NVCTM, and KATE. 
\end{abstract}


\section{Introduction}

Understanding large collections of unstructured text remains a persistent problem. Unsupervised models offer a formalism for exposing a collection’s themes and have been used to aid information retrieval \cite{wei2006lda}, discover patterns in the medical data sets \cite{goudarzvand2019early, goudarzvand2018analyzing}, understand authorship in the texts \cite{jafariakinabad2019syntactic}, mining news media \cite{zolnoorimining}, cyberbullying detection \cite{samghabadi2020detecting}, video frame prediction \cite{hosseini2019inception} and time series forecasting \cite{hosseini2020direct}. Topic models have also been applied outside text to learn natural scene categories in computer vision \cite{fei2005bayesian}; and understand the connection between Bayesian models and cognition \cite{griffiths2007unifying}.

An autoencoder is a neural network that learns data representations by reconstructing the input data at the output layer (i.e., $y(i) = x(i)$)  \cite{bengio2007greedy, goodfellow2016deep, lecun2015deep}. Autoencoders learn the most salient features of the input data by constraining part of the hidden layers, called the bottleneck, often by reducing its dimension less than the input layer. Consequently, the bottleneck neurons become the learned features. While autoencoders have been successfully used in several applications such as denoising images \cite{vincent2010stacked}, conventional autoencoders face several challenges when used for textual data, including the text’s high-dimensionality and sparsity. Moreover, autoencoders are known to learn trivial representations of textual data due to its power-law word distribution \cite{zhai2016semisupervised}. Therefore, several approaches have emerged to address the challenges mentioned above, such as deep belief nets for topic modeling \cite{maaloe2015deep} and neural variational inference for text processing \cite{miao2016neural}. Nonetheless, such methods face accuracy reduction when focused on achieving better topics. 

\begin{figure}
\includegraphics[width=\columnwidth]{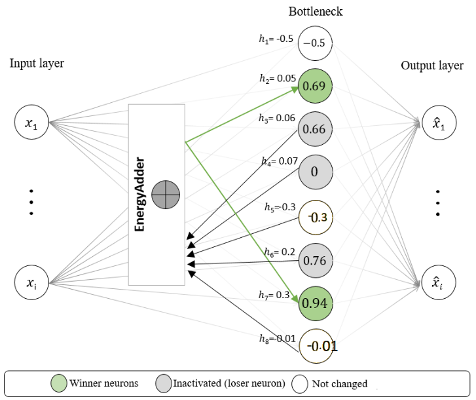}
\centering
\caption{An example illustrating the SCAT approach. Layers are fully connected.}\label{fig:Appraoch}
\end{figure}

In this paper, we introduce a k-competitive autoencoder for extracting meaningful representative features with notable accuracy results. While the k-competitive learning approach was used before in several autoencoders, including K-Spare \cite{makhzani2013k} and KATE \cite{chen2017kate}; these methods differ in the competition criteria. For example, K-Sparse aims at enforcing sparsity in the hidden layers by keeping $k$ highest activities in the training phase and $\alpha k$ highest activities in the testing phase ($k$ and $\alpha$ are hyperparameters). The k-competitive approach of KATE selects $k$ winner neurons composed of $k/2$ largest positive activations and $k/2$ largest absolute negative activations. Then, the $k$ winner neurons gain the energy (i.e., activation value) of the loser neurons. However, through our extensive experiments, we observed that some important words/topics are often represented in neurons with small positive activation values, which are ignored by conventional k-competitive autoencoders. Thus, they never get the chance to be represented in the final model. We also observed that neurons with negative activation values have very small to no effect on the model performance when included in competition selection criteria. 

In contrast, the novelty of our k-competitive learning approach, SCAT, stems from proposing a “fair” competition by providing a second chance for the smallest positive neurons to reveal their potential, i.e., important topics that otherwise are ignored—and hence the name Second Chance Autoencoder (SCAT). Our approach selects the $k/2$ largest (strongest) positive activations and $k/2$ smallest (weakest) positive activations as the $k$ winners, which then gain the energy of the loser neurons. Note that $k$ is a hyperparameter that represents the number of neurons to be included in the competition, and it strongly correlates to the number of topics. Our experiments suggest that setting $k=\# topics/2$ yields higher performance results. 
Our qualitative and quantitative experiments prove that SCAT achieves close to or better than the current state-of-the-art performance on three datasets (20 Newsgroups \cite{lang1995newsweeder}, Reuters \cite{lewis2004rcv1}, and Wiki10+ \cite{zubiaga2012enhancing}) across several tasks compared to LDA \cite{blei2003latent}, K-Sparse, NVCTM \cite{liu2019neural}, and KATE.

\section{Approach}
SCAT, illustrated in Fig. \ref{fig:Appraoch}, is a competitive learning approach that not only supports the competition among the largest activation values, similar to conventional k-competitive approaches, but grants a \textit{second chance} to the neurons with lowest non-negative activation values. The underlying idea was drawn from our observations that some essential features are often buried in neurons with weak non-negative activation values. We also noticed that the inclusion of the weakest negative neurons in the competition process, similar to KATE's approach, does not yield much improvement in the overall process compared to the inclusion of the weakest positive neurons. Consequently, and triggered by the positive impact of second chances in life, our approach grants the weakest non-negative neurons a second chance to prove that their values might have a potential to correspond to distinctive features.

In particular, the SCAT training cycle includes the following steps (refer to Algorithms 1 and 2). First, a feedforward step is carried out to the bottleneck layer, which then selects the winner neurons and assigns them the energy of the loser neurons. Energy, the sum of the activation values of the loser neurons, is reassigned to the winner neurons equally. Loser neurons are then inactivated (i.e., set to zero).  Note that neurons with negative values do not participate in the competition process. They are left unchanged due to their insignificant contribution. Then, we use weight tying to initialize the weights of the decoder part (hidden-to-output layers). We use the sigmoid activation function at the output layer. Finally, during backpropagation, the gradients will flow through the winner neurons and ignore the loser neurons since they were inactivated.  It is important to mention here that our algorithm does not require any special steps for encoding inference cases after training the model since the network is already well-trained to represent distinctive features. 

Figure 1 illustrates an example of the training process in SCAT. For simplicity, the example includes only eight neurons in the bottleneck layer, and thus we set $k = 8/4 = 2$.  The original activation values computed by the feedforward step before energy redistribution are shown at the left side of the neurons. The winner neurons are colored in Green, which include (1) the strongest neuron, $h_7=0.3$, and (2) the weakest non-negative neuron, $h_2=0.05$. Then, the energy from the loser neurons, excluding the negative ones, are added up, i.e., $E = 0.01+0.2+0.3+0.07+0.06 = 0.64$, and reassigned to the winner neurons. The rest of the neurons are not included in the learning process (loser neurons are inactivated, and negative neurons are not changed). In Fig. \ref{fig:Appraoch}, the activation values resulted from the SCAT layer are shown inside the neurons.

\begin{algorithm}[t]
\SetKwProg{Pn}{procedure}{:}{}
  \Pn{Training Phase}{
\For{e in epochs}{
$z = tanh(Wx+b)$

$\hat{z} = sscat\_layer(k, z)$

$\hat{x} = sigmoid(W^T\hat{z}+c)$

$loss = cross\_entropy(x, \hat{x})$

$backpropagate\_error$
}}
\caption{The Training Phase}
\label{Algmain}
\end{algorithm}

\begin{algorithm}[t]
\SetKwProg{Pn}{function}{:}{\KwRet $\hat{z}$}
  \Pn{scat\_layer($k$, $z$)}{
$z_l$=[largest $\ceil{k/2}$ activations in $z$]
  
$z_s$=[smallest $\floor{k/2}$ activations in $z$]

\tcp{compute total energy}
$E$ = $\sum_{i \in \{z: z= z-[z_l, z_s], z >0 \}} z_i$

\tcp{reallocate energy to winners}
$z_l += E$

$z_s += E$

\tcp{inactivate loser neurons}

\For{$z$ in $z$ - $[z_l, z_s]$}
{
z = 0}

$\hat{z}$=[$z_l$, $z_s$, $z$]
}
\caption{SCAT Layer Definition}
\label{alg:SCAT}
\end{algorithm}

\section{Results}
We compare the results of our SCAT model to the following models: (1) LDA: a probabilistic topic model that uses the bag-of-words technique to model a topic and a mixture of topics to model a document. (2) K-Sparse: an autoencoder that enforces sparsity in the hidden layers by keeping $k$ highest activities in the training phase and $k\alpha$ highest activities in the testing phase. K-Sparse uses linear activation functions, while the non-linearity in the model derives from the selection of $k$ highest activities. (3) NVCTM: a novel model that proposes the idea of centralized transformation flow to capture the correlations among topics by reshaping topic distributions. The implementation of this model is not available, so we compared our results to the results reported in their paper \cite{liu2019neural}. (4) KATE: a shallow autoencoder model with a competitive hidden layer that selects $k$ strongest positive neurons and weakest negative neurons. KATE also requires an additional hyperparameter, $\alpha$, to amplify the energy value.

\subsection{Quantitative Analysis}
\textit{Document Classification:}
The classification experiment included training a simple softmax multi-class classifier with a cross-entropy loss function on the 20 Newsgroups dataset. The classification precision, recall, and F1 scores are listed under the 20 Newsgroups column in Table \ref{tab:comparision1}. We set the number of topics to 50. It is obvious from the table that competition based autoencoders achieve better results than conventional models, LDA. KATE achieves 70\% for all three measurements outperforming NVCTM, K-Sparse, and LDA. However, our SCAT autoencoder outperforms all models achieving 73\% scores on all three measurements.  

\begin{table}
\centering
\caption{Comparison of the Performance Evaluation on 20 Newsgroup}
\label{tab:comparision1}
\resizebox{0.70\columnwidth}{!}{%
\begin{tabular}{@{}cccc@{}}
\toprule
Model         & Precision     & Recall        & F1 score      \\ \midrule
LDA           & 0.42          & 0.50          & 0.46          \\
K-Sparse      & 0.42          & 0.42          & 0.42          \\
NVCTM         & 0.57          & 0.56          & 0.57          \\
KATE          & 0.70          & 0.70          & 0.70          \\
\textbf{SCAT} & \textbf{0.73} & \textbf{0.73} & \textbf{0.73} \\ \bottomrule
\end{tabular}%
}
\end{table}

\subsection{Qualitative Analysis}
We illustrate that our model can learn more semantically meaningful representations from textual data compared to the above-mentioned baseline models using the 20 Newsgroups dataset, with the number of topics set to 20--matching the number of classes. The results are listed in Table \ref{tab:comparisionII}. We observe from the table that SCAT model generates the most semantically meaningful topics. For example, in the Religion category, the topics \textit{god}, \textit{bible}, \textit{christ}, and \textit{heaven} are all strongly related to Religion. In the Sport category, words like \textit{players}, \textit{hockey}, \textit{game}, \textit{league}, and \textit{season} illustrate the most meaningful representations among the rest of the words generated by the other models. 

\section{Related work}
K-Sparse \cite{makhzani2013k} aims at enforcing sparsity in the hidden layers by keeping the $k$ highest activities in the training phase and the $\alpha$ highest activities in the testing phase ($k$ and $\alpha$ are hyperparameters). K-Sparse uses linear activation functions for the hidden neurons, while the non-linearity in the model derives from the selection of the $k$ highest activities. K-Spars achieved better classification results than denoising autoencoders, models trained with dropout, and Restricted Boltzmann Machines when applied for textual data. 

The authors of \cite{zhai2016semisupervised} developed a semi-supervised autoencoder and a loss function to overcome the scalability challenges of high text dimensionality. Their proposed model and loss function significantly improved the classification results on sentiment analysis applications. Another important work in this area includes KATE (K-competitive Autoencoder for TExt), which was proposed in \cite{chen2017kate}. KATE builds on top of K-Spars and aims at learning meaningful representations by introducing competition among the neurons of the hidden layers. Particularly, $k$ neurons with the strongest positive and absolute negative activation values gain the power of the rest of neurons; and, thus, they become specialized in learning more meaningful representations. The k-strongest neurons (both positive and negative) are referred to as winners while the rest of the neurons are referred to as losers. KATE has proven to improve the state-of-the-art results of document classification over variational, contractive, and K-Spars autoencoders. 

A recent related work, NVCTM (Neural Variational Correlated Topic Modeling), was proposed in \cite{liu2019neural}. NVCTM introduced the idea of centralized transformation flow to capture the correlations among topics by reshaping topic distributions. It consists of two components: the inference network with a centralized transformation flow and a multinomial softmax generative model. The extensive experiments of NVCTM validated its efficiency in capturing perplexity, topic coherence, and document classification tasks.

\begin{table}
\caption{Comparison of Selected Topics from the 20 Newsgroup}
\label{tab:comparisionII}
\resizebox{\columnwidth}{!}{%
\begin{tabular}{|c|c|c|}
\hline
Category               & Moldel         & Topics                                                                                                                         \\ \hline
\multirow{4}{*}{Religion} &
  LDA &
  \begin{tabular}[c]{@{}c@{}}god, people, jesus, Christian, subject,\\  bible, church, christ, time, life\end{tabular} \\ \cline{2-3} 
                       & K-Sparse      & \begin{tabular}[c]{@{}c@{}}god, world, people, origin, subject, \\ pad, christian, bottom, application, mind\end{tabular}      \\ \cline{2-3} 
                       & Kate          & \begin{tabular}[c]{@{}c@{}}god, michael, rutgers, dod, jesus, \\ christian, bije, drive, uga, christ\end{tabular}              \\ \cline{2-3} 
                       & \textbf{SCAT} & \textbf{\begin{tabular}[c]{@{}c@{}}god, bible, athos, gordon, food, \\ geb, rutgers, heaven, christ, disease\end{tabular}}     \\ \hline
\multirow{4}{*}{Politics} &
  LDA &
  \begin{tabular}[c]{@{}c@{}}armenian, turkish, armenians, people, war, \\ turkey, muslim, muslims, armenia, turks\end{tabular} \\ \cline{2-3} 
                       & K-Sparse      & \begin{tabular}[c]{@{}c@{}}israel, ca, card, system, israeli, \\ national, government, state, armerican, lawfound\end{tabular} \\ \cline{2-3} 
                       & KATE          & \begin{tabular}[c]{@{}c@{}}article, writes, nsa, gov, news, \\ org, israel, israeli, university, jews\end{tabular}             \\ \cline{2-3} 
 &
  \textbf{SCAT} &
  \textbf{\begin{tabular}[c]{@{}c@{}}government, population, members, gun, space, \\ united, states, citizesn, political, congress\end{tabular}} \\ \hline
\multirow{4}{*}{Sport} & LDA           & \begin{tabular}[c]{@{}c@{}}game, team, year, subject, games, \\ hockey, players, play, writes, good\end{tabular}               \\ \cline{2-3} 
                       & K-Sparse      & \begin{tabular}[c]{@{}c@{}}team, steve, good, mile, players, \\ season, hockey, internet, win, article\end{tabular}            \\ \cline{2-3} 
                       & KATE          & \begin{tabular}[c]{@{}c@{}}game, games, hockey, team, red, \\ win, season, nhl, play, leafs\end{tabular}                       \\ \cline{2-3} 
                       & \textbf{SCAT} & \textbf{\begin{tabular}[c]{@{}c@{}}players, game, teams, key, league, \\ season, proposal, year, hockey\end{tabular}}          \\ \hline
\end{tabular}%
}
\end{table}

\section{Conclusions }
We proposed a novel autoencoder named SCAT,  Second Chance Autoencoder for Text. The underlying approach of SCAT relies upon the idea of k-competitive learning, in which $k$ winner neurons participate in the learning process and gain the power of the loser neurons, which then become inactivated. Our experiments validated that our approach achieves very close or better performance results on document classification and provides more semantically meaningful topics compared to the baselines models. Our experiments and model training and validation tasks were managed using ModelKB \cite{gharibi2019modelkb, gharibi2019automated}. 

\section{Future Work}
Our future work aims at reporting more comprehensive experiments using additional datasets and baseline models. We also aim to introduce an enhanced version of SCAT that uses more sophisticated competition criteria that further enhances the autoencoder results. 

\printbibliography
\end{document}